\documentclass{article}
\usepackage{spconf}
\usepackage{graphicx}
\usepackage[normalem]{ulem}
\usepackage{amsmath,amssymb,amsfonts}
\usepackage{bm} 
\usepackage{multirow}
\usepackage{colortbl}
\usepackage{cite}
\usepackage{algorithmic}
\usepackage{graphicx}
\usepackage{textcomp}
\usepackage{xcolor}
\usepackage{multirow}

\def\BibTeX{{\rm B\kern-.05em{\sc i\kern-.025em b}\kern-.08em
    T\kern-.1667em\lower.7ex\hbox{E}\kern-.125emX}}
\usepackage{tikz}

\title{S3I-PointHop: SO(3)-Invariant PointHop for 3D Point Cloud Classification}
\name{Pranav~Kadam \textsuperscript{1}, 
Hardik~Prajapati \textsuperscript{1}, 
Min~Zhang \textsuperscript{1}, 
Jintang Xue \textsuperscript{1}, 
Shan Liu \textsuperscript{2},
C.-C.~Jay~Kuo \textsuperscript{1}
\thanks{The authors acknowledge the gift support from the Tencent Media
Lab as well as the Center for Advanced Research Computing (CARC) at the
University of Southern California for providing computing resources that
have contributed to the research results reported within this
publication. URL: https://carc.usc.edu.}}

\address{University of Southern California, Los Angeles, California, USA$^1$\\
Tencent Media Lab, Palo Alto, California, USA$^2$}

\begin{document}

\maketitle
\ninept

\begin{abstract}

Many point cloud classification methods are developed under the
assumption that all point clouds in the dataset are well aligned with
the canonical axes so that the 3D Cartesian point coordinates can be
employed to learn features. When input point clouds are not aligned, the
classification performance drops significantly. In this work, we focus
on a mathematically transparent point cloud classification method called
PointHop, analyze its reason for failure due to pose variations, and
solve the problem by replacing its pose dependent modules with rotation
invariant counterparts. The proposed method is named SO(3)-Invariant
PointHop (or S3I-PointHop in short). We also significantly simplify the
PointHop pipeline using only one single hop along with multiple spatial
aggregation techniques. The idea of exploiting more spatial information
is novel. Experiments on the ModelNet40 dataset demonstrate the
superiority of S3I-PointHop over traditional PointHop-like methods. 

\end{abstract}


\section{Introduction}

Due to numerous applications in autonomous vehicles and robotics
perception, immersive media processing, 3D graphics, etc., 3D Point
Clouds have emerged to be a popular form of representation for 3D vision
tasks. Research and development on point cloud data processing has
attracted a lot of attention. Recent trends show a heavy inclination
towards development of learning-based methods for point clouds. 

One of the primary tasks in point cloud understanding is object
classification. The task is to assign a category label to a 3D point
cloud object scan. The unordered nature of 3D point cloud demands
methods to be invariant to $N!$ point permutations for a scan of $N$
points. It was demonstrated in the pioneering work called PointNet
\cite{qi2017pointnet} that permutation invariance can be achieved using
a symmetric function such as the maximum value of point feature
responses. Besides permutations, invariance with respect to rotations is
desirable in many applications such as 3D registration.  In particular,
point cloud features are invariant with any 3D transformation in the
SO(3) group; namely, the group of $3\times3$ orthogonal matrices
representing rotations in 3D. 

Achieving rotation invariance can guarantee that point clouds expressed
in different orientations are regarded the same and, thereby, the
classification result is unaffected by the pose. State-of-the-art
methods do not account for rotations, and they perform poorly in
classifying different rotated instances of the same object. In most
cases, objects are aligned in a canonical pose before being fed into a
learner. Several approaches have been proposed to deal with this
problem. First is data augmentation, where different rotated instances
of the same object are presented to a learner. Then, the learner
implicitly learns to reduce the error metric in classifying similar
objects with different poses. This approach leads to an increase in the
computation cost and system complexity. Yet, there is no guarantee in
rotational invariance. A more elegant way is to design point cloud
representations that are invariant to rotations. Thus, point cloud
objects expressed in different orientations are indistinguishable to
classifiers. Another class of methods are based on SO(3) equivariant
networks, where invariance is obtained as a byproduct of the equivariant
point cloud features. 

Point cloud classification based on the green learning principle was
first introduced in PointHop \cite{zhang2020pointhop}. The work is
characterized by its mathematical transparency and lightweight nature.
The methodology has been successfully applied to point cloud
segmentation \cite{zhang2020unsupervised, zhang2022gsip} and
registration \cite{kadam2022r}.  For point cloud classification, both
PointHop and its follow-up work PointHop++ \cite{zhang2020pointhop++}
assume that the objects are pre-aligned. Due to this assumption, these
methods fail when classifying objects with different poses. 

In this work, we propose an SO(3) invariance member for the PointHop
family, and name it S3I-PointHop. This is achieved through the
derivation of invariant representations by leveraging principal
components, rotation invariant local/global features, and point-based
eigen features.  Our work has two main contributions. First, the pose
dependent octant partitioning operation in PointHop is replaced by an
ensemble of three rotation invariant representations to guarantee SO(3)
invariance. Second, by exploiting the rich spatial information, we
simplify multi-hop learning in PointHop to one-hop learning in
S3I-PointHop. Specifically, two novel aggregation schemes (i.e., conical
and spherical aggregations in local regions) are proposed, which makes
one-hop learning possible. 

The rest of this paper is organized as follows. Related material is
reviewed in Sec. \ref{sec:review}. The S3I-PointHop method is proposed
in Sec. \ref{sec:method}. Experimental results are presented in Sec.
\ref{sec:experiments}. Finally, concluding remarks are given in Sec.
\ref{sec:conclusion}. 

%%%%%%%%%%%%%%%%%%%%%%%%%%%%%%%%%%%%%%%%%%%%%%%%%%%%%%%
\begin{figure*}[htb]
\centerline{\includegraphics[width=7in]{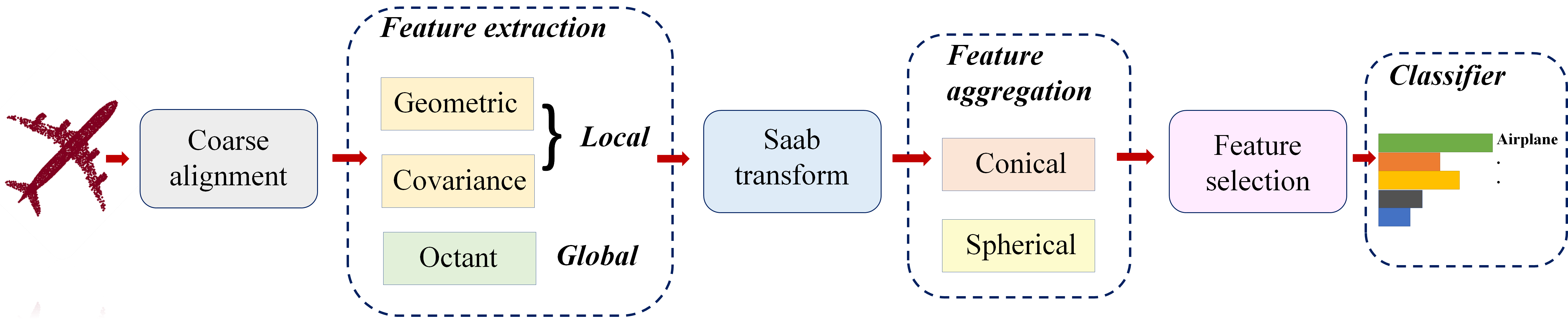}}
\caption{An overview of the proposed S3I-PointHop method: 1) an input point
cloud scan is approximately aligned with the principal axes, 2) local and
global point features are extracted and concatenated followed by the
Saab transform, 3) point features are aggregated from different conical
and spherical volumes, 4) discriminant features are selected using DFT
and a linear classifier is used to predict the object class.}
\label{fig:architecture}
\end{figure*}
%%%%%%%%%%%%%%%%%%%%%%%%%%%%%%%%%%%%%%%%%%%%%%%%%%%%%%%%

\section{Related Work} \label{sec:review}

\subsection{Green Point Cloud Learning}

Green learning (GL) \cite{kuo2022green} is a data-driven learning
methodology. It uses training data statistics to derive representations
without labels. The learning process utilizes the Saab transform
\cite{kuo2019interpretable} or the channel-wise Saab transform
\cite{chen2020pixelhop++}. GL is a radical departure from neural
networks. It has achieved impressive results for point cloud data
processing. For example, PointHop and PointHop++ offer competitive
performance in classification of aligned point clouds. They both have
three main modules: 1) hierarchical attribute construction based on the
distribution of neighboring points in the 3D space and attribute
dimensionality reduction using the Saab/channel-wise Saab transform, 2)
feature aggregation, and 3) classification.  The capability of GL has
been demonstrated by follow-ups, including point cloud segmentation
\cite{zhang2020unsupervised, zhang2022gsip}, registration
\cite{kadam2020unsupervised, kadam2022r}, odometry
\cite{kadam2022greenpco}, and pose estimation \cite{kadam2022pcrp}.
GL-based point cloud processing techniques are summarized in
\cite{liu20213d}.  Similar to early point cloud classification methods,
PointHop and PointHop++ fail to classify objects of arbitrary poses.

\subsection{Rotation Invariant Networks}

Early pioneering deep networks for point cloud processing tasks such as
PointNet \cite{qi2017pointnet}, PointNet++ \cite{qi2017pointnet++},
DGCNN \cite{wang2019dynamic} and PointCNN \cite{li2018pointcnn} are
susceptible to point cloud rotations. Designing rotation invariant
networks has been popular for 3D registration when global alignment is
needed. Methods such as PPFNet \cite{deng2018ppfnet} and PPF-FoldNet
\cite{deng2018ppf} achieve partial and full invariance to 3D
transformations, respectively. The idea behind any rotation invariant
method is to design a representation that is free from the pose
information. This is done by exploiting properties of 3D transformations
such as preservation of distances, relative angles, and principal
components.  Global and local rotation invariant features for
classification were proposed in \cite{li2021rotation}, which form a
basis of our method. Ambiguities associated with global PCA alignment
were analyzed and a disambiguation network was proposed in
\cite{li2021closer}.  Another approach is the design of equivariant
neural networks that achieve invariance via certain pooling operations.
SO(3)- and SE(3)-equivariant convolutions make networks equivariant to
the 3D rotation and 3D roto-translation groups, respectively. Exemplary
work includes the Vector Neurons \cite{deng2021vector} for
classification and segmentation, results in \cite{chen2021equivariant,
li2021leveraging} for category-level pose estimation. 

\section{Proposed S3I-PointHop Method} \label{sec:method}

The S3I-PointHop method assigns a class label to a point cloud scan,
$X$, whose points are expressed in an arbitrary coordinate system. Its
block diagram is shown in Fig. \ref{fig:architecture}. It comprises of
object coarse alignment, feature extraction, dimensionality reduction,
feature aggregation, feature selection and classification steps as
detailed below.

\subsection{Pose Dependency in PointHop}

The first step in PointHop is to construct a 24-dimensional local
descriptor for every point based on the distribution of 3D coordinates
of the nearest neighbors of that point. 3D rotations are distance
preserving transforms and, hence, the distance between any two points
remains the same before and after rotation. As a consequence, the
nearest neighbors of points are unaffected by the object pose. However,
the use of 3D coordinates makes PointHop sensitive to rotations since
the 3D Cartesian coordinates of every point change with rotation.
Furthermore, the 3D space surrounding the current point is partitioned
into 8 octants using the standard coordinate axes. The coordinate axes
change under different orientations of the point cloud scan. 
We align an object with its three principal axes. The PCA
alignment only offers a coarse alignment, and it comes with several
ambiguities as pointed out in \cite{li2021closer}. Furthermore, object
asymmetries may disturb the alignment since PCA does not contain
semantic information. Yet, fine alignment is not demanded.  Here, we
develop rotation invariant features based on PCA aligned objects. 

\subsection{Feature Extraction}

Local and global information fusion is effective in feature
learning for point cloud classification \cite{wang2019dynamic}.  To
boost the performance of S3I-PointHop, three complementary features are
ensembled. The first feature set contains the omni-directional octant
features of points in the 3D space as introduced in PointHop. That is,
the 3D space is partitioned into eight octants centered at each point as
the origin. The mean of 3D coordinates of points in each octant then
constitute the 24D octant feature.  The second feature set is composed
by eigen features \cite{hackel2016fast} obtained from the covariance
analysis of the neighborhood of a point.  They are functions of the
three eigen values derived from the Singular Value Decomposition (SVD)
of the local covariance matrix. The 8 eigen features comprise of
linearity, planarity, anisotropy, sphericity, omnivariance, verticality,
surface variation and eigen entropy. They represent the surface
information in the local neighborhood. The third feature set is formed
by geometric features derived from distances and angles in local
neighborhoods as proposed in \cite{li2021rotation}. For simplicity, we
replace the geometric median in \cite{li2021rotation} with the mean of
the neighboring coordinates.  The 12D feature representation is found
using the $K$ nearest neighbors, leading to a pointwise $12 \times K$
matrix.  While a small network is trained in \cite{li2021rotation} to
aggregate these features into a single vector, we perform a channel-wise
max, mean and $l_2$-norm pooling to yield a 36D vector of local
geometric feature.  The octant, covariance and geometric features are
concatenated together to build a 68D ($24+8+36=68)$ feature vector.
After that, the Saab transform is performed for dimension reduction. 

%%%%%%%%%%%%%%%%%%%%%%%%%%%%%%%%%%%%%%%%%%%%%%%%%%%%%%%
\begin{figure}[htb]
\centerline{\includegraphics[width=2.5in]{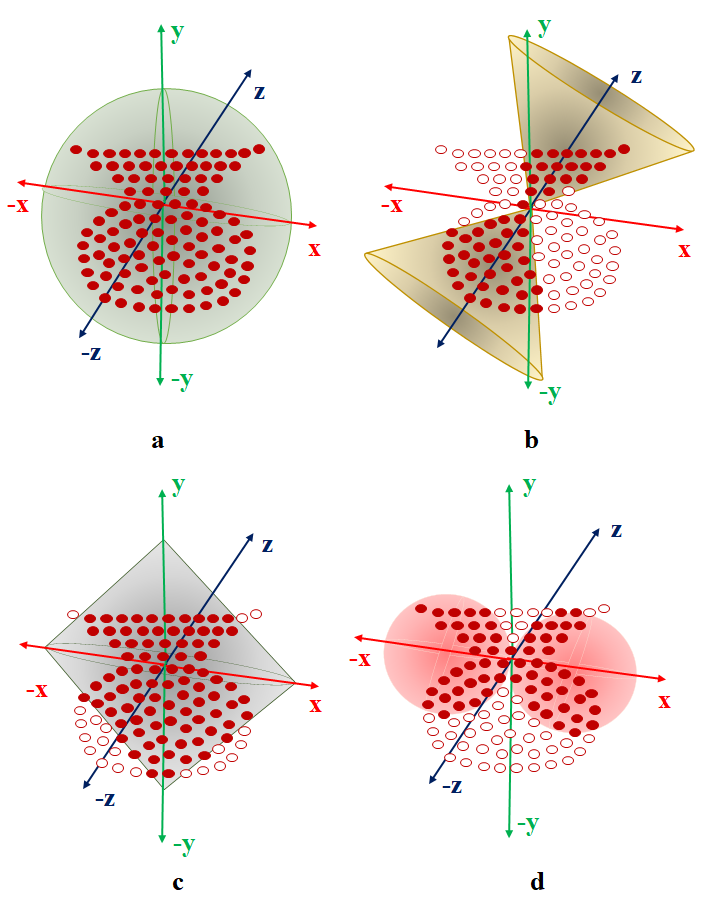}}
\caption{Illustration of conical and spherical aggregation. The
conventional ``global pooling" is shown in (a), where features of all
points are aggregated at once. The proposed ``regional pooling" schemes
are depicted in (b)-(d), where points are aggregated only in distinct
spatial regions. Only, the solid red points are aggregated. For better
visual representation, cones/spheres along only one axis are shown.  (b)
and (c) use the conical pooling while (d) adopts spherical pooling in
local regions.} \label{fig:aggregation}
\end{figure}
%%%%%%%%%%%%%%%%%%%%%%%%%%%%%%%%%%%%%%%%%%%%%%%%%%%%%%%%

\subsection{Feature Aggregation}

The point features need to be aggregated into a global point cloud
feature for classification. A symmetric aggregation function such as max
or average pooling is a popular choice for feature aggregation.  Four
aggregations (the max, mean, $l_1$ norm, and $l_2$ norm) have been used
in PointHop and PointHop++.  Instead of aggregating all points globally
at once as shown in Fig.  \ref{fig:aggregation} (a), we propose to
aggregate subsets of points from different spatial regions here. We
consider regions of the 3D volume defined by cones and spheres. 

For conical aggregation, we consider two types of cones, one with tip at
the origin and the other with tip at a unit distance along the principal
axes. They are illustrated in Figs.  \ref{fig:aggregation} (b) and (c),
respectively.  The latter cone cuts the plane formed by the other two
principal axes in a unit circle and vice versa for the former. For each
principal axis, we get four such cones, two along the positive axis and
two along the negative. Thus, 12 cones are formed for all three axes in
total. For each cone, only the features of points lying inside the cone
are pooled together.  The pooling methods are the max, mean, variance,
$l_1$ norm, and $l_2$ norm. This means for a single point feature dimension, we get a 5D feature vector from each cone.

\begin{figure}[htb]
\centerline{\includegraphics[width=3.4in]{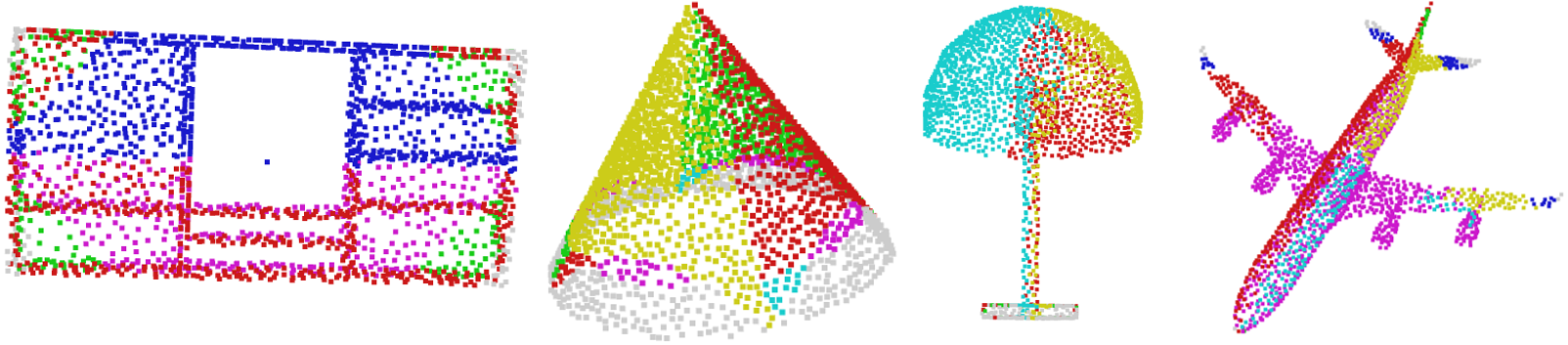}}
\caption{An example of conical aggregation. For every point cloud object,
points lying in each cone are colored uniquely.}\label{fig:cones}
\end{figure}

For spherical aggregation, we consider four spheres of a quarter radius
centered at a distance of positive/negative one and three quarters from
the origin along each principal axis.  One example is illustrated in
Fig. \ref{fig:aggregation} (d).  This gives 12 spheres in total.  Points
lying in each sphere are pooled together in a similar manner as cones.
For instance, points lying in different cones for four point cloud
objects are shaded in Fig. \ref{fig:cones}. 

Unlike max/average pooling, aggregating local feature descriptors into a
global shape descriptor such as Bag of Words (BoW) or Vector of Locally
Aggregated Descriptors (VLAD) \cite{jegou2010aggregating} is common in
traditional literature.  On the other hand, the region-based local
spatial aggregation has never been explored before. These resulting
features are powerful in capturing local geometrical characteristics of
objects. 

%%%%%%%%%%%%%%%%%%%%%%%%%%%%%%%%%%%%%%%%%%%%%%%%%%%%%%%
%%%%%%%%%%%%%%%%%%%%%%%%%%%%%%%%%%%%%%%%%%%%%%%%%%%%%%%%

\subsection{Discriminant Feature Selection and Classification}

In order to select a subset of discriminant features for classification,
we adopt the Discriminant Feature Test (DFT) as proposed in
\cite{yang2022supervised}.  DFT is a supervised learning method that can
rank features in the feature space based on their discriminant power.
Since they are calculated independently of each other, the DFT
computation can be parallelized. Each 1D feature $f^i$ of all point
clouds are collected and the interval $[f^i_{min},f^i_{max}]$ is
partitioned into two subspaces $S^i_L$ and $S^i_R$ about an optimal
threshold $f^i_{op}$. Then, the purity of each subspace is measured by a
weighted entropy loss function. A smaller loss indicates stronger
discriminant power. DFT helps control the number of features fed to the
classifier.  As shown in Sec. \ref{sec:experiments}, it improves the
classification accuracy significantly and prevents classifier from
overfitting. In our experiments, we select top 2700 features. Finally,
we train a linear least squares classifier to predict the object class. 

\section{Experiments} \label{sec:experiments}

We evaluate the proposed S3I-PointHop method for the point cloud
classification task on the ModelNet40 dataset \cite{wu20153d}, which
consists of 40 object classes.  Objects in ModelNet40 are pre-aligned.
We rotate them in the train and test sets in the following experiments.
The rotation angles are uniformly sampled in $[0,2\pi]$. We use $z$ to
denote random rotations along the azimuthal axis and $SO(3)$ to indicate
rotations about all three orthogonal axes. In Tables \ref{tab:pointhop}, \ref{tab:others} and \ref{tab:ablation_study}, $z/SO(3)$ means that the training set follows the
$z$ rotations while the test set adopts $SO(3)$ rotations, and so on.
For all experiments, we set the numbers of nearest neighbors in
calculating geometric, covariance, and octant features to be 128, 32,
and 64, respectively. 

\subsection{Comparison with PointHop-family Methods}

Table \ref{tab:pointhop} compares the performance of S3I-PointHop,
PointHop \cite{zhang2020pointhop}, PointHop++ \cite{zhang2020pointhop++}
and R-PointHop \cite{kadam2022r}. Clearly, S3I-PointHop outperforms the
three benchmarking methods by a huge margin.  Although R-PointHop was
proposed for point cloud registration and not classification, we include
it here due to its rotation invariant feature characteristics.  Similar
to the global aggregation in PointHop and PointHop++, we aggregate the
point features of R-PointHop and train a Least Squares classifier. We
also report the classification accuracy with only one hop for these
methods. Both PointHop and PointHop++ perform poor since their features
are not invariant to rotations. Especially, for the $z/SO(3)$ case,
there is an imbalance in the train and test sets, the accuracy is worse.
R-PointHop only considers local octant features with respect to a local
reference frame. Although they are invariant to rotations, they are not
optimal for classification. 

%%%%%%%%%%%%%%%%%%%%%%%%%%%%%%%%%%%%%%%%%%%%%%%%%%%%%%%%%%%%%%%%%%%%
\begin{table}[htbp]
\centering
\caption{Classification accuracy comparison of PointHop-family methods.} \label{tab:pointhop}
\renewcommand\arraystretch{1.3}
\newcommand{\tabincell}[2]{\begin{tabular}{@{}#1@{}}#2\end{tabular}}
\resizebox{\columnwidth}{!}{
\begin{tabular}{c | c | c |  c | c } \hline 
Method & \# hops & z/z & z/SO(3) & SO(3)/SO(3) \\ \hline
 \multirow{ 2}{*}{PointHop \cite{zhang2020pointhop}} & 1 & 70.50  & 21.35 & 45.70  \\ 
 & 4 & 75.12  & 22.85 & 50.48  \\ \hline
\multirow{ 2}{*}{PointHop++ \cite{zhang2020pointhop++}}   & 1 & 9.11  & 7.90 & 9.09  \\ 
 & 4 & 82.49 & 20.62 & 57.61  \\ \hline
\multirow{ 2}{*}{R-PointHop \cite{kadam2022r}} & 1 & 53.44  & 53.42 & 53.44 \\ 
 & 4 & 64.87  & 64.86 & 64.86   \\ \hline
S3I-PointHop & 1 & \bf{83.10} & \bf{83.10} & \bf{83.10}   \\ \hline
\end{tabular}
}
\end{table}
%%%%%%%%%%%%%%%%%%%%%%%%%%%%%%%%%%%%%%%%%%%%%%%%%%%%%%%%%%%%%%%%%%%%

\subsection{Comparison with Deep Learning Networks}

We compare the performance of S3I-PointHop with 4 deep-learning-based
point cloud classification networks in Table \ref{tab:others}. They are
PointNet \cite{qi2017pointnet}, PointNet++ \cite{qi2017pointnet++},
PointCNN \cite{li2018pointcnn} and Dynamic Graph CNN (DGCNN)
\cite{wang2019dynamic}. Since these methods were originally developed
for aligned point clouds, we retrain them with rotated point clouds and
report the corresponding results. We see from the table that
S3I-PointHop outperforms these benchmarking methods significantly. These
methods offer reasonable accuracy when rotations are restricted about
the azimuthal (z) axis. However, they are worse when rotations are
applied about all three axes. 

%%%%%%%%%%%%%%%%%%%%%%%%%%%%%%%%%%%%%%%%%%%%%%%%%%%%%%%%%%%%%%%%%%%%
\begin{table}[htbp]
\centering
\caption{Comparison with Deep Learning Networks.} \label{tab:others}
\renewcommand\arraystretch{1.3}
\newcommand{\tabincell}[2]{\begin{tabular}{@{}#1@{}}#2\end{tabular}}
% \resizebox{\columnwidth}{!}{
\begin{tabular}{c | c |  c | c } \hline 
Method & z/z & z/SO(3) & SO(3)/SO(3) \\ \hline
 PointNet \cite{qi2017pointnet} & 70.50  & 21.35 & 45.70  \\ \hline
 PointNet++ \cite{qi2017pointnet++} & 75.12  & 22.85 & 50.48  \\ \hline
PointCNN  \cite{li2018pointcnn} & 82.11  & 24.89 & 51.66  \\ \hline
 DGCNN \cite{wang2019dynamic} & 82.49 & 20.62 & 57.61  \\ \hline

 S3I-PointHop & \bf{83.10}  & \bf{83.10} & \bf{83.10}   \\ \hline
\end{tabular}
% }
\end{table}
%%%%%%%%%%%%%%%%%%%%%%%%%%%%%%%%%%%%%%%%%%%%%%%%%%%%%%%%%%%%%%%%%%%%

\subsection{Ablation Study}

It is worthwhile to consider the contributions of different elements in
S3I-PointHop. To do so, we conduct an ablation study and report the
results in Table \ref{tab:ablation_study}. From the first three rows, it
is evident that the global octant features are most important, and their
removal results in the highest drop in accuracy. The results also
reinforce the fact that locally oriented features such as those in
R-PointHop are not optimal for classification. In rows 4 and 5, we
compare the proposed spatial aggregation scheme (termed as local
aggregation) with global pooling as done in PointHop. The accuracy
sharply drops by 12\% when only the global aggregation is used. Clearly,
global aggregation is not appropriate in S3I-PointHop. Finally, we show
in the last row that the accuracy drops to 78.56\% without DFT.  The is
because, when the feature dimension is too high, the classifier can
overfit easily without DFT. 

%%%%%%%%%%%%%%%%%%%%%%%%%%%%%%%%%%%%%%%%%%%%%%%%%%%%%%%%%%%%%%%%%%%%
\begin{table}[htbp]
\centering
\caption{Ablation Study} \label{tab:ablation_study}
\renewcommand\arraystretch{1.3}
\newcommand{\tabincell}[2]{\begin{tabular}{@{}#1@{}}#2\end{tabular}}
\resizebox{\columnwidth}{!}{
\begin{tabular}{c  c   c | c c |c | c} \hline 

% \multicolumn{3}{c|}{\multirow{ 2}{*}{Feature}} & \multicolumn{2}{c|}{\multirow{ 2}{*}{Aggregation}} & \multirow{ 2}{*}{DFT} &  \multirow{ 2}{*}{Test}  \\ \hline

\multicolumn{3}{c|}{Feature} & \multicolumn{2}{c|}{Aggregation} & \multirow{ 2}{*}{DFT} & \multirow{ 2}{*}{SO(3)/SO(3)}  \\ 

\tabincell{c}{Geometric}  &  \tabincell{c}{Covariance}  
& \tabincell{c}{Octant} & \tabincell{c}{Local}  &  \tabincell{c}{Global} & &   \\ \hline 

& \checkmark & \checkmark & \checkmark & & \checkmark & 82.49 \\ \hline 
\checkmark & & \checkmark & \checkmark & & \checkmark & 82.45 \\ \hline 
\checkmark & \checkmark & & \checkmark & & \checkmark & 80.75 \\ \hline 
\checkmark & \checkmark & \checkmark & \checkmark & & \checkmark & \bf{83.10} \\ \hline 
\checkmark & \checkmark & \checkmark & & \checkmark & \checkmark & 71.02 \\ \hline 
\checkmark & \checkmark & \checkmark & \checkmark & & & 78.56 \\ \hline 
\end{tabular}
}
\end{table}
%%%%%%%%%%%%%%%%%%%%%%%%%%%%%%%%%%%%%%%%%%%%%%%%%%%%%%%%%%%%%%%%%%%%

\subsection{Discussion}

One advantage of S3I-PointHop is that its rotation invariant
features allow it to handle point cloud data captured from different
orientations. To further support this claim, we retrain PointHop with
PCA coarse alignment as a pre-processing step during the training and
the testing. The test accuracy is 78.16\% and 74.10\% with four-hop and
one-hop, respectively.  This reinforces that only the PCA alignment is
not the reason for the performance gain of S3I-PointHop.  While efforts
to learn rotation invariant features were already made in R-PointHop, we
see that the lack of global features in it degrades its performance. On
the other hand, appending the same global feature to R-PointHop does not
help in the registration problem. 

An interesting aspect of S3I-PointHop is its use of a single hop (rather
than four hops such as in PointHop). It is generally perceived that
deeper networks perform better than shallower counterparts.  However,
the use of multiple spatial aggregations on top of a single hop,
S3I-PointHop can achieve good performance. This leads to the benefit of
reducing the training time and the model size as explained below.

In any point-cloud processingn method, one of the most costly operations
is the nearest neighbor search. To search the $k$ nearest neighbors of
each of $N$ points, the complexity of an efficient algorithm is $O(k
\log N)$.  PointHop uses the nearest neighbor search in four hops and
three intermediate farthest point downsampling operations. In contrast,
the nearest neighbor search is only conducted once for each point in
S3I-PointHop. Another costly operation is the PCA in the Saab transform.
It is performed only once in S3I-PointHop.  Its model size is 900 kB,
where only one-hop Saab filters are stored. 

\section{Conclusion and Future Work} \label{sec:conclusion}

A point cloud classification method called S3I-PointHop was proposed.
It extends PointHop-like methods for 3D classification of objects which
have arbitrary orientations. S3I-PointHop extracts local and global
point neighborhood information using an ensemble of geometric,
covariance and octant features. Only a single hop is adopted in
S3I-PointHop followed by conical and spherical aggregations of point
features from multiple spatial regions.  There are several possible
extensions of this work.  It is desired to further improve the
performance of S3I-PointHop and compare it with that of state-of-the-art
rotation invariant and equivariant networks.  Furthermore, it is
interesting to examine the application of single-hop rotation invariant
methods to the registration problem and the pose estimation problem. 

\bibliographystyle{IEEEtran}
\bibliography{refs}

\end{document}